\documentclass[11pt,letterpaper]{article}
\usepackage{emnlp2017}
\usepackage{times}
\usepackage{latexsym}

\usepackage{caption}
\usepackage{url}

\usepackage{xcolor}
\usepackage{listings}
\colorlet{light-gray}{gray!20}
\usepackage{graphicx}
\usepackage{array}

\newcommand\blfootnote[1]{%
  \begingroup
  \renewcommand\thefootnote{}\footnote{#1}%
  \addtocounter{footnote}{-1}%
  \endgroup
}

\emnlpfinalcopy

\def\emnlppaperid{***}

\title{NeuroNER: an easy-to-use program for \\named-entity recognition based on neural networks}
\author{Franck Dernoncourt\thanks{\hspace{3mm}These authors contributed equally to this work.}\\
	    MIT\\
	    {\tt francky@mit.edu}
	   \And
	 Ji Young Lee\footnotemark[1]\\
   	MIT\\
   {\tt jjylee@mit.edu}
   \And
	 Peter Szolovits \\
   	MIT\\
   {\tt psz@mit.edu}
   }

\date{}

\begin{document}

\maketitle

\vspace{-0.6cm}
\begin{abstract}\vspace{-0.2cm}
Named-entity recognition (NER) aims at identifying entities of interest in a text. Artificial neural networks (ANNs) have recently been shown to outperform existing NER systems. However, ANNs remain challenging to use for non-expert users. In this paper, we present NeuroNER, an easy-to-use named-entity recognition tool based on ANNs.  Users can annotate entities using a graphical web-based user interface (BRAT): the annotations are then used to train an ANN, which in turn predict entities' locations and categories in new texts. NeuroNER  makes this annotation-training-prediction flow smooth and accessible to anyone.
\end{abstract}

\vspace{-0.2cm}
\section{Introduction}
\vspace{-0.2cm}
Named-entity recognition (NER) aims at identifying entities of interest in the text, such as location, organization and temporal expression. Identified entities can be used in various downstream applications such as patient note de-identification and information extraction systems. They can also be used as features for machine learning systems for other natural language processing tasks.

Early systems for NER relied on rules defined by humans. Rule-based systems are time-consuming to develop,
and cannot be easily transferred to new types of texts or entities. To address these issues, researchers have developed machine-learning-based algorithms for NER, using a variety of learning approaches, such as fully supervised learning, semi-supervised learning, unsupervised learning, and active learning. 
NeuroNER is based on a fully supervised learning algorithm, which is the most studied approach~\cite{nadeau2007survey}.

Fully supervised approaches to NER include support vector machines (SVM)~\cite{asahara2003japanese}, maximum entropy models~\cite{borthwick1998nyu}, decision trees~\cite{sekine1998nyu} as well as sequential tagging methods such as hidden Markov models~\cite{bikel1997nymble}, Markov maximum entropy models~\cite{kumar2006named}, and conditional random fields (CRFs)~\cite{mccallum2003early,tsai2006nerbio,benajiba2008arabic,filannino2013mantime}.
Similar to rule-based systems, these approaches rely on handcrafted features, which are challenging and time-consuming to develop and may not generalize well to new datasets.

More recently, artificial neural networks (ANNs) have been shown to outperform other supervised algorithms for NER~\cite{collobert2011natural,lample2016neural,lee2016feature,labeau-loser-allauzen:2015:EMNLP,dernoncourt2016identification}. The effectiveness of ANNs can be attributed to their ability to learn effective features jointly with model parameters directly from the training dataset, instead of relying on handcrafted features developed from a specific dataset. However, ANNs remain challenging to use for non-expert users.

\vspace{-0.0cm}

\paragraph{Contributions}
NeuroNER makes state-of-the-art named-entity recognition based on ANN available to anyone, by focusing on usability. To enable users to create or modify annotations for a new or existing corpus, NeuroNER interfaces with the web-based annotation program BRAT~\cite{stenetorp2012brat}. NeuroNER makes the annotation-training-prediction flow smooth and accessible to anyone, while leveraging the state-of-the-art prediction capabilities of ANNs. NeuroNER is open source and freely available online\footnote{NeuroNER can be found online at:}\blfootnote{\hspace{0cm}\url{http://neuroner.com}}.

\section{Related Work}

Existing publicly available NER systems geared toward non-experts do not use ANNs. For example, Stanford NER~\cite{finkel2005incorporating}, ABNER~\cite{settles.bioinf05}, the MITRE Identification Scrubber Toolkit  (MIST)~\cite{aberdeen2010mitre},~\cite{boag2015cliner}, BANNER~\cite{leaman2008banner} and NERsuite~\cite{cho2010nersuite} rely on CRFs. GAPSCORE uses SVMs~\cite{chang2004gapscore}.  Apache cTAKES~\cite{savova2010mayo} and Gate's ANNIE~\cite{cunningham1996gate,maynard2003multilingual} use mostly rules. NeuroNER, the first ANN-based NER system for non-experts, is more generalizable to new corpus due to the ANNs' capability to learn effective features jointly with model parameters.

Furthermore, in many cases, the NER systems assume that the user already has an annotated corpus formatted in a specific data format. As a result, users often have to connect their annotation tool with the NER systems by reformatting annotated data, which can be time-consuming and error-prone.  Moreover, if users want to manually improve the  annotations predicted by the NER system (e.g., if they use the NER system to accelerate the human annotations), they have to perform additional data conversion. NeuroNER streamlines this process by incorporating BRAT, a widely-used and easy-to-use annotation tool.

\begin{figure*}[!ht]
  \centering

  \includegraphics[width=\textwidth]{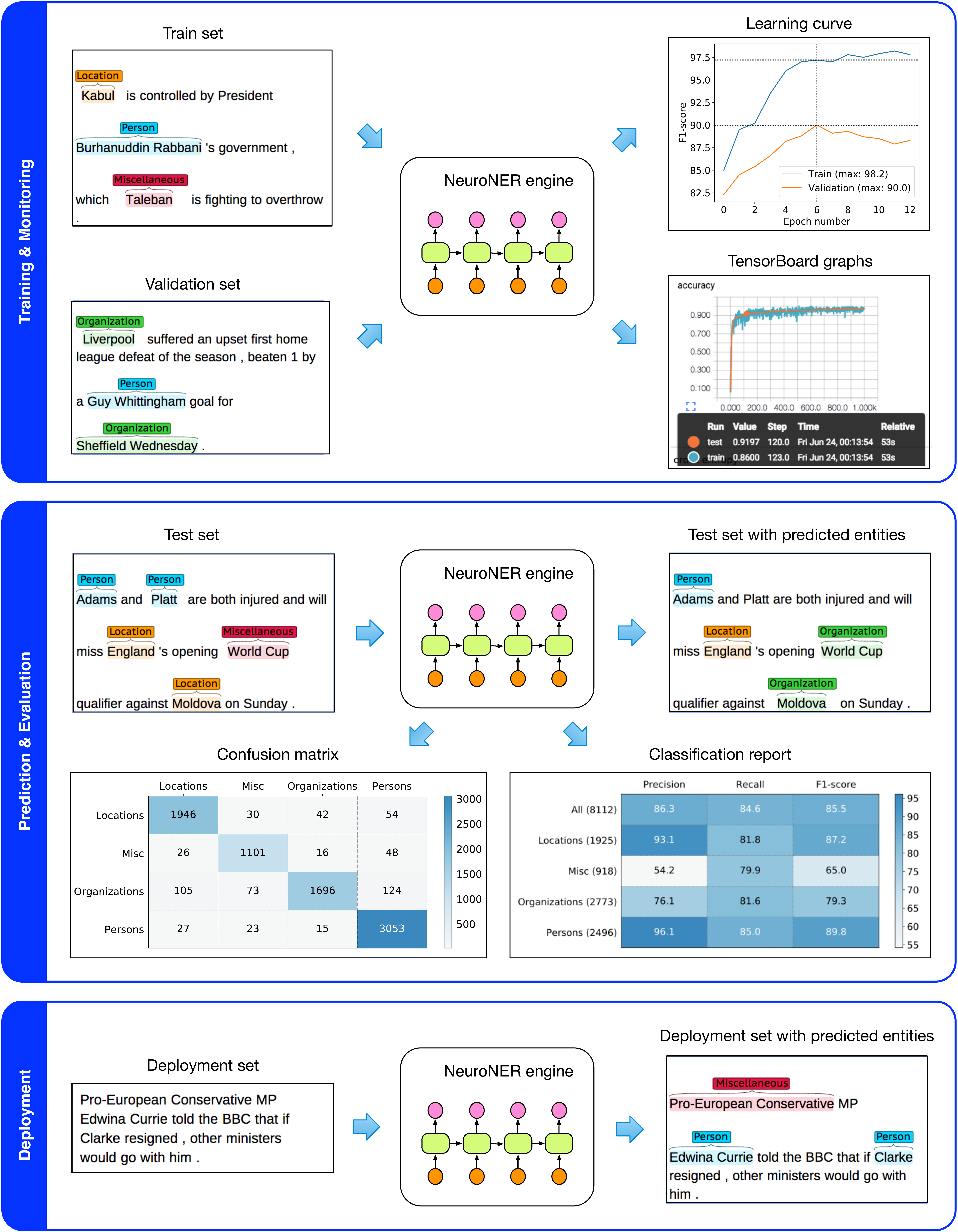}
  \vspace{-0.6cm}
  \caption{NeuroNER system overview. In the NeuroNER engine, the training set is used to train the parameters of the ANN, and the validation set is used to determine when to stop training. The user can monitor the training process in real time via the learning curve and TensorBoard. To evaluate the trained ANN, the labels are predicted for the test set: the performance metrics can be calculated and plotted by comparing the predicted labels with the gold labels. The evaluation can be done at the same time as the training if the test set is provided along with the training and validation sets, or separately after the training or using a pre-trained model. Lastly, the NeuroNER engine can label the deployment set, i.e. any new text without gold labels.}
  \label{fig:model}
\end{figure*}

\section{System Description}
NeuroNER comprises two main components: an NER engine and an interface with BRAT. 
NeuroNER also comes with real-time monitoring tools for training, and pre-trained models that can be loaded to the NER engine in case the user does not have access to labelled training data.
Figure~\ref{fig:model} presents an overview of the system.

\vspace{0.0cm}
\subsection{NER engine}
The NER engine takes as input three sets of data with gold labels: the training set, the validation set, and the test set. Additionally, it can also take as input the deployment set, which refers to any new text without gold labels that the user wishes to label.
The files that comprise each set of data should be in the same format as used for the annotation tool BRAT or the CoNLL-2003 NER shared task dataset~\cite{tjong2003introduction}, and organized in the corresponding folder.
\vspace{0.4cm}

The NER engine's ANN contains three layers:
\begin{itemize}
\setlength\itemsep{0.1em}
\item Character-enhanced token-embedding layer,
\item Label prediction layer,
\item Label sequence optimization layer.
\end{itemize}
\vspace{.0cm}

The character-enhanced token-embedding layer maps each token to a vector representation. The sequence of vector representations corresponding to a sequence of tokens is then input to label prediction layer, which outputs the sequence of vectors containing the probability of each label for each corresponding token.  Lastly, the label sequence optimization layer outputs the most likely sequence of predicted labels based on the sequence of probability vectors from the previous layer. All layers are learned jointly.

The ANN as well as the training process have several hyperparameters such as  character embedding dimension, character-based token-embedding LSTM dimension,
token embedding dimension, and dropout probability. All hyperparameters may be  specified in a configuration file that is human-readable, so that the user does not have to dive into any code. Listing~\ref{ConfigParser} presents an excerpt of  the  configuration file.

\vspace{.4cm}

\begin{lstlisting}[frame=single, backgroundcolor=\color{light-gray}, basicstyle=\footnotesize\ttfamily, language=Java, numbers=none, numberstyle=\tiny\color{black},caption= {Excerpt of  the  configuration file used to define the ANN as well as the training process.
Only the  \texttt{dataset\_folder} parameter needs to be changed by the user: the other parameters have reasonable default values, which the user may optionally tune.},captionpos=b,label={ConfigParser}]
[dataset]
dataset_folder               = dat/conll

[character_lstm]
using_character_lstm         = True
char_embedding_dimension     = 25
char_lstm_dimension          = 50

[token_lstm]
token_emb_pretrained_file    = glove.txt
token_embedding_dimension    = 200
token_lstm_dimension         = 300

[crf]
using_crf                    = True
random_initial_transitions   = True

[training]
dropout                      = 0.5
patience                     = 10
maximum_number_of_epochs     = 100
maximum_training_time        = 10 
number_of_cpu_threads        = 8
\end{lstlisting}

\subsection{Real-time monitoring for training}

As training an ANN may take many hours, or even a few days on very large datasets, NeuroNER provides the user with real-time feedback during the training  for monitoring purpose. 
Feedback is given through two different means: plots generated by NeuroNER, and TensorBoard.

\paragraph{Plots} NeuroNER generates several plots showing the training progress and outcome at each epoch. Plots include the evolution of the overall F1-score over time, confusion matrices visualizing the number of correct versus incorrect predictions for each class, and classification reports showing the F1-score, precision and recall for each class.
\paragraph{TensorBoard} As NeuroNER is based on TensorFlow
, it leverages the functionalities of TensorBoard. TensorBoard is a suite of web applications for inspecting and understanding TensorFlow runs and graphs. It allows to view in real time the performances achieved by the ANN being trained. Moreover, since it is web-based, these performances can be conveniently shared with anyone remotely. Lastly, since graphs generated by TensorBoard are interactive, the user may gain further insights on the ANN performances.

\subsection{Pre-trained models}

Some users may prefer not to train any ANN model, either due to time constraints or unavailable gold labels. 
For example, if the user wants to tag protected health information, they might not be able to have access to a labeled identifiable dataset. 
To address this need, NeuroNER provides a set of pre-trained models. Users are encouraged to contribute by uploading their own trained models. NeuroNER also comes with several pre-trained token embeddings, either with word2vec~\cite{mikolov2013efficient,mikolov2013distributed,mikolov2013linguistic} or GloVe~\cite{pennington2014glove}, which the NeuroNER engine can load easily once specified in the configuration file.

\subsection{Annotations}

NeuroNER is designed to smoothly integrate with the freely available web-based annotation tool BRAT, so that non-expert users may create or improve annotations. Specifically, NeuroNER addresses two main use cases:
\vspace{-0.2cm}
\begin{itemize}
\item creating new annotations from scratch, e.g. if the goal is to annotate a dataset for which no gold label is available,
\item improving the annotations of an already labeled dataset: the annotations may have been done by another human or by a previous run of NeuroNER. 
\end{itemize}
In the latter case, the user may use NeuroNER interactively, by iterating between manually improving the annotations and running the NeuroNER engine with the new annotations to obtain more accurate annotations.

NeuroNER can take as input datasets in the BRAT format, and outputs BRAT-formatted predictions, which makes it easy to start training directly from the annotations as well as visualize and analyze the predictions.
We chose BRAT for two main reasons: it is easy to use, and it can be deployed as a web application, which allows crowdsourcing. As a result, the user may quickly gather a vast amount of annotations by using crowdsourcing marketplaces such as Amazon Mechanical Turk~\cite{buhrmester2011amazon} and CrowdFlower~\cite{finin2010annotating}.

\subsection{System requirements}

NeuroNER runs on Linux, Mac OS X, and Microsoft Windows. It requires Python 3.5, TensorFlow 1.0~\cite{abadi2016tensorflow}, scikit-learn ~\cite{pedregosa2011scikit}, and BRAT.
 A setup script is provided to make the installation straightforward. It can use the GPU if available, and the number of CPU threads and GPUs to use can be specified in the configuration file.

\subsection{Performances}

\begin{table} [b]
\footnotesize
\centering
\setlength{\extrarowheight}{3pt}
\setlength{\arraycolsep}{5pt}
\begin{tabular}{|l|c|c|c|c|c|c|}
\hline
\textbf{Model} & CoNLL 2003 & i2b2 2014  \\
 \hline
Best published		& 90.9			&	97.9	 \\ 
NeuroNER		& 90.5			&	97.7	 \\ 
\hline
\end{tabular}
\caption{F1-scores (\%) on the test set comparing NeuroNER with the best published methods in the literature, viz.~\cite{passos2014lexicon} for CoNLL 2003, \cite{dernoncourt2016identification} for i2b2 2014.} \label{tab:result-comparisons}
\end{table}

To assess the quality of NeuroNER's predictions, we use two publicly and freely available datasets for named-entity recognition: CoNLL 2003 and i2b2 2014. CoNLL 2003~\cite{tjong2003introduction} is a widely studied dataset with 4 usual types of entity: persons, organizations, locations  and miscellaneous names.
We use the English version.

The i2b2 2014 dataset~\cite{stubbs2015automated} was released as part of the 2014 i2b2/UTHealth shared task Track 1. It is the largest publicly available dataset for de-identification, which is a form of named-entity recognition where the entities are protected health information such as patients' names and patients' phone numbers. 22~systems were submitted for this shared task.

Table~\ref{tab:result-comparisons} compares NeuroNER with state-of-the-art systems on CoNLL 2003 and i2b2 2014. Although the hyperparameters of NeuroNER were not optimized for these datasets (the default hyperparameters were used), the performances  of NeuroNER  are on par with the state-of-the-art systems.

\section{Conclusions}

In this article we have presented NeuroNER, an ANN-based NER tool that is accessible to non-expert users and yields state-of-the-art results. Addressing the need of many users who want to create or improve annotations, NeuroNER smoothly integrates with the web-based annotation tool BRAT.

\bibliography{references}
\bibliographystyle{emnlp_natbib}

\appendix

\end{document}